\documentclass{article}
\usepackage[numbers]{natbib}

\usepackage[final]{funasr_tech_report}

\usepackage[utf8]{inputenc} %
\usepackage[T1]{fontenc}    %
\usepackage{hyperref}       %
\usepackage{url}            %
\usepackage{booktabs}       %
\usepackage{amsfonts}       %
\usepackage{authblk} %
\usepackage{adjustbox}      %
\usepackage{nicefrac}       %
\usepackage{microtype}      %
\usepackage{xcolor}         %
\usepackage{graphicx}
\usepackage{subcaption} %
\usepackage{amsfonts}       %
\usepackage{amsmath}        %
\usepackage{cite}
\usepackage{multicol}
\usepackage{multirow}
\usepackage{enumitem}
\usepackage{pifont}
\usepackage{array}
\usepackage{caption}
\usepackage[table]{xcolor}

\hypersetup{
    colorlinks=true,
    linkcolor=black,
    citecolor=black,
    urlcolor=QwenPurple
}

\usepackage{fontawesome5}

\definecolor{QwenPurple}{HTML}{7357D9}
\definecolor{QwenPurple}{HTML}{6F4FD8}
\definecolor{QwenPurpleLight}{HTML}{EEE9FF}
\definecolor{QwenLavender}{HTML}{F3F0FC}
\definecolor{ErrorRed}{HTML}{B54747}

\newcolumntype{L}[1]{%
  >{\raggedright\arraybackslash}p{#1}%
}

\newcommand{\bad}[1]{%
  \textcolor{ErrorRed}{\bfseries #1}%
}

\newcommand{\good}[1]{%
  \textcolor{QwenPurple}{\bfseries #1}%
}

\newcolumntype{V}[1]{%
  >{\raggedright\arraybackslash}m{#1}%
}

\usepackage[utf8]{inputenc}
\usepackage[T1]{fontenc}
\usepackage{CJKutf8}

\title{Qwen-Audio-3.0-ASR Technical Report}

\author{Qwen-Audio ASR Team}
\affil{
  $\vcenter{\hbox{\includegraphics[height=1.0em]{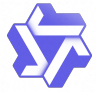}}}$
  \textbf{Alibaba Token Foundry}
  \\
  \vspace{2mm}
 \href{https://qwenaudio.github.io/qwen-audio-3.0-asr/}{
  \faGlobe\enspace\textbf{Project Page}
}
  }

\begin{document}

\maketitle

\begin{abstract}
In recent years, automatic speech recognition (ASR) has witnessed transformative advancements driven by three complementary paradigms: data scaling, model scaling, and deep integration with large language models (LLMs). However, bridging the gap between academic benchmark performance and real-world production utility remains a persistent challenge, particularly in handling diverse regional dialects, dynamic entities and hotwords, long-range contextual information, and disfluent spontaneous speech. In this report, we present \textbf{Qwen-Audio-3.0-ASR}, a Mixture-of-Experts (MoE) LLM-based ASR system designed to address these production demands through a unified, instruction-following framework. The model is built upon the Qwen backbone, and is trained on tens of millions of hours of large-scale speech data. Qwen-Audio-3.0-ASR supports transcription across 30 languages and 16 Chinese dialectal varieties spanning eight major dialect regions. Beyond multilingual and dialectal recognition, the model provides production-oriented capabilities including industry-domain entity recognition, hierarchical hotword customization, native single-pass transcription polishing, and long-audio contextual modeling. We further develop a dedicated streaming variant, \textbf{Qwen-Audio-3.0-ASR-Streaming}, for latency-sensitive applications. Extensive evaluations on Chinese, English, multilingual, and real-world industrial test sets demonstrate state-of-the-art or highly competitive recognition performance across a broad range of evaluation conditions, with strong performance relative to leading commercial and proprietary systems including GPT-4o Transcribe and Gemini 3.1 Pro.

\end{abstract}

\section{Introduction}
\label{sec:introduction}

\begin{figure}[h]
    \centering
    \includegraphics[width=1.0\linewidth]{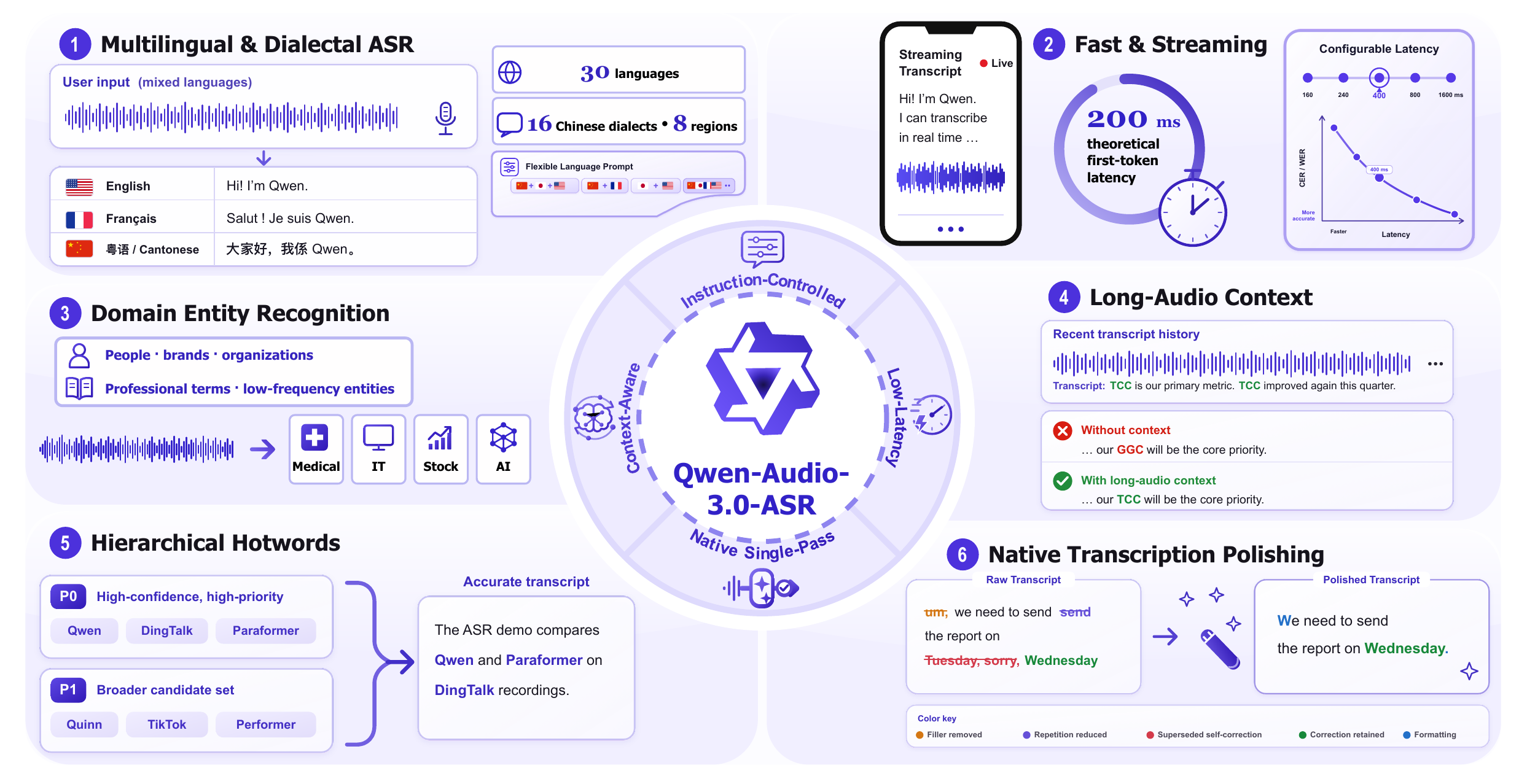} 
    \caption{\textbf{Overview of Qwen-Audio-3.0-ASR and its production-oriented capabilities.} The system supports multilingual and dialectal recognition, low-latency streaming, domain entity recognition, long-audio contextual modeling, hierarchical hotword customization, and native single-pass transcription polishing. The central ring highlights four system-level properties: instruction-controlled decoding, context awareness, native single-pass generation, and low latency.}
    \label{fig:overview}
\end{figure}

Automatic speech recognition (ASR) is a task of converting spoken language into written text. It is a foundational technology underpinning a wide range of real-world applications from voice assistants and live captioning to meeting transcription, intelligent customer service, and cross-lingual communication. As these applications expand, the demands placed on ASR systems have grown far beyond simply converting speech to text: users now expect accurate, context-aware, and immediately readable transcription across diverse languages, accents, and acoustic environments.

To meet these growing expectations, three complementary paradigms have driven ASR from traditional neural-network-based pipelines to modern large-model-based architectures in recent years: \textbf{data scaling}, \textbf{model size scaling}, and \textbf{deep integration with large language models (LLMs)}. Data scaling has proven to be a fundamental driver. For example, the Whisper series~\citep{DBLP:conf/icml/RadfordKXBMS23} demonstrated that expanding training data from 3K to over 680K hours yields more than a 20-point WER reduction, and the most competitive systems today leverage datasets covering tens of millions of hours. Model size scaling amplifies these gains: within the same Whisper family, increasing parameters from 38M to 1.5B produced over 40 points of additional multilingual WER improvement. The third paradigm, LLM integration, represents a methodological shift: by harnessing the linguistic knowledge and contextual reasoning of frontier LLMs, models such as Seed-ASR~\citep{DBLP:journals/corr/abs-2407-04675} and FireRedAudio~\citep{li2026fireredaudio} resolve semantic ambiguities and generate more coherent transcriptions than acoustic-only systems.

Recent studies have continued to extend LLM-based ASR along several production-relevant directions. For multilingual recognition, Mixture-of-Experts (MoE) projectors with dynamic downsampling have been shown to improve cross-lingual adaptation and speech--text alignment \citep{lin2026multilingualmoe}, while MEUSLI connects a pretrained speech encoder to open multilingual LLMs and supports ASR across 28 European languages \citep{concina2026meusli}. For latency-sensitive applications, Uni-ASR explores a unified LLM-based architecture for both streaming and non-streaming recognition \citep{xia2026uniasr}. For context awareness and hotword recognition, recent approaches have explored large-vocabulary hotword retrieval with reinforcement-learning-based adaptation \citep{kong2025hotwordrl}, as well as acoustic cues and bias-word position prediction for rare-word recognition \citep{novitasari2026contextualbiasing}.

However, bridging the gap between academic benchmark success and real-world production utility remains a persistent challenge. Conventional ASR models often simultaneously struggle with three practical demands: (1) handling diverse regional dialects within a single model rather than deploying per-market checkpoints; (2) performing robust contextual recognition of dynamic entities and hotwords that evolve faster than any fixed training set; and (3) delivering clean, reader-friendly transcripts free of disfluencies and self-corrections, without an expensive post-hoc rewriting pass.

In this report, we present \textbf{Qwen-Audio-3.0-ASR}, a Mixture-of-Experts LLM-based ASR (LLM-ASR) system that addresses the three production demands through a unified, instruction-following framework, as illustrated in Figure~\ref{fig:overview}. Qwen-Audio-3.0-ASR is the successor to Fun-ASR~\citep{an2025fun}. Compared with its predecessor, this generation replaces the previous 7B dense language-model decoder with the Qwen MoE backbone, and further unifies multilingual and dialectal recognition, contextual conditioning, hotword customization, and transcription polishing within a flexible instruction-controlled interface. Relative to major commercial and proprietary ASR systems evaluated in this report, Qwen-Audio-3.0-ASR is distinguished by jointly supporting these production-oriented capabilities within a unified recognition framework. In addition, we develop a dedicated low-latency streaming variant, Qwen-Audio-3.0-ASR-Streaming, for latency-sensitive applications. Trained on large-scale data spanning tens of millions of hours, the system exhibits the following key capabilities:

\begin{itemize}[leftmargin=*,noitemsep]
    \item \textbf{Flexible multilingual and dialectal coverage.} Qwen-Audio-3.0-ASR takes a list of target languages as part of its input instruction and adapts decoding to the specified language combination at inference time. Through this single instruction-following interface, the system covers 30 languages and 16 dialectal varieties spanning eight major Chinese dialect regions.
    \item \textbf{Industry-domain entity recognition and hierarchical hotword customization.} A continuously expanding entity-mining and data-synthesis pipeline systematically improves recognition of domain-specific terminology across industries. This is complemented by a hierarchical hotword mechanism that separates high-confidence, contextually critical terms (P0) from a broader candidate pool (P1), using an LLM to perform semantic classification and assign the tiers.
    \item \textbf{Single-pass disfluency polishing and long-context modeling.} Rather than relying on a cascaded ASR-then-LLM-rewrite pipeline, Qwen-Audio-3.0-ASR performs disfluency removal, self-correction resolution, and punctuation normalization within the same decoding pass, conditioned on an instruction flag to eliminate the additional inference latency of an external LLM-rewriting stage. Long-context conditioning on previously recognized text further suppresses cross-utterance entity inconsistencies (e.g., in meetings and lectures) without requiring a manually maintained hotword list.
    \item \textbf{Ultra-low-latency streaming.} The streaming variant, \textbf{Qwen-Audio-3.0-ASR-Streaming}, is designed for latency-sensitive production scenarios. It sustains ultra-low first-token latency while delivering recognition accuracy that substantially outpaces competing industrial streaming engines across both Chinese and English.
    \item \textbf{State-of-the-art performance.} Extensive evaluations across Chinese, English, and multilingual benchmarks, as well as real-world industrial test sets, demonstrate the strong and well-balanced performance of Qwen-Audio-3.0-ASR. On Chinese and English benchmarks, the model achieves state-of-the-art results on several widely used test sets and remains highly competitive on the others. On multilingual benchmarks, it achieves the best macro-average error rates on GigaSpeechBench and Common Voice 15 among the evaluated systems, while maintaining competitive performance on FLEURS. Overall, Qwen-Audio-3.0-ASR consistently demonstrates strong performance against both open-source systems and leading commercial and proprietary models.
\end{itemize}

This report is organized as follows. Section~\ref{sec:architecture} introduces the model architecture. Section~\ref{sec:training} elaborates on the training and reinforcement learning paradigms and the corresponding  training data. Deployment is shown in Section~\ref{sec:deployment}. Evaluation experiments are presented in Section~\ref{sec:exp}.

\section{Model Architecture}
\label{sec:architecture}

\begin{figure}[h]
    \centering
    \includegraphics[width=1.0\linewidth]{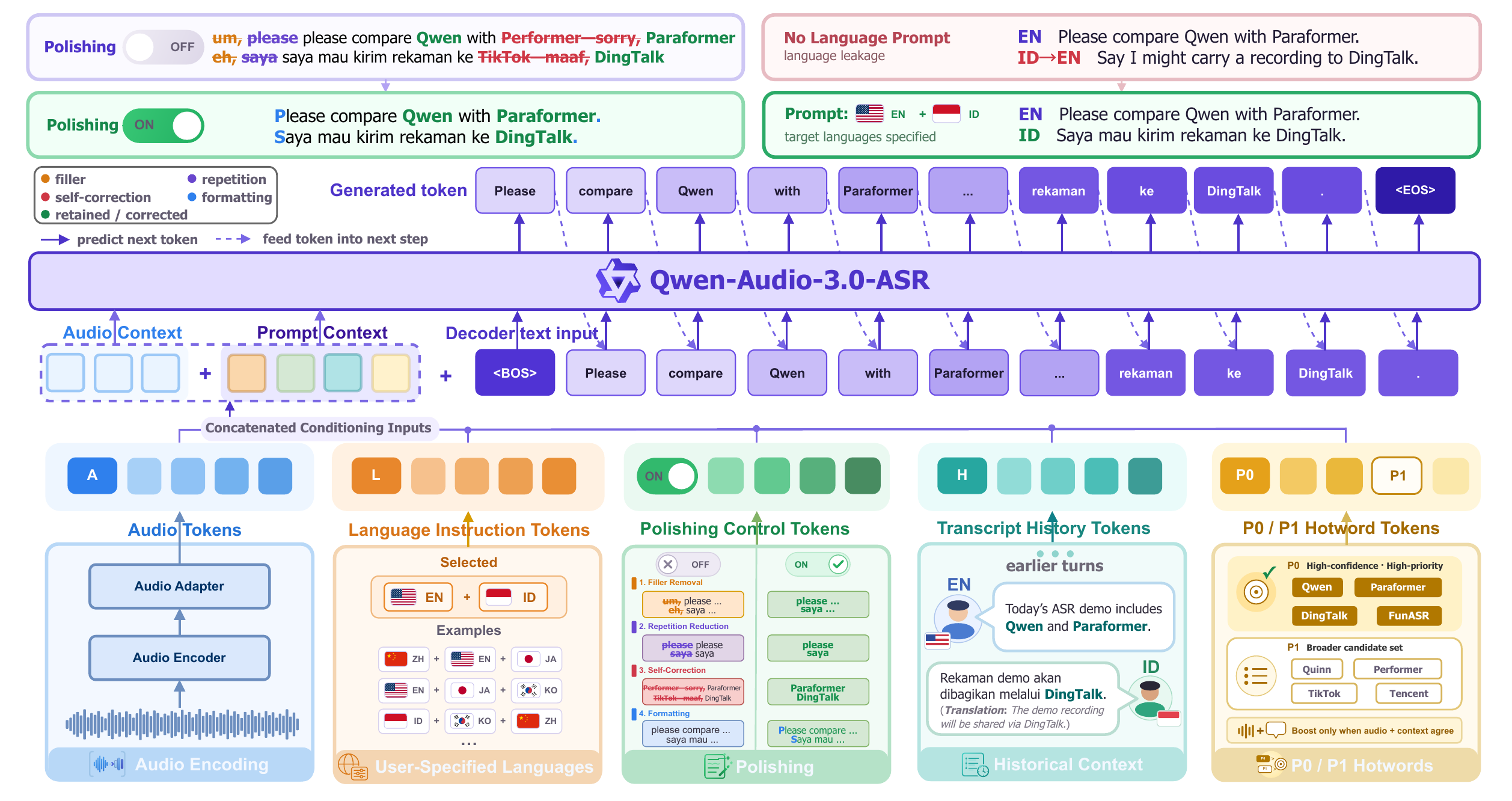} 
    \caption{Overview of the instruction-controlled decoding workflow of Qwen-Audio-3.0-ASR.}
    \label{fig:model}
\end{figure}

Figure~\ref{fig:model} illustrates the overall decoding workflow of Qwen-Audio-3.0-ASR, showing how audio inputs and user-specified conditions, including target languages, historical context, polishing instructions, and P0/P1 hotwords, are jointly incorporated into the decoding process.
Qwen-Audio-3.0-ASR retains the overall topology of the previous-generation Fun-ASR model~\citep{an2025fun}, comprising four key components: (1) an audio encoder, implemented as multiple Transformer encoder layers, that extracts representations from the input speech; (2) an audio adapter, implemented as a two-layer Transformer encoder, that connects the audio encoder output to the LLM; (3) a CTC decoder built on top of the audio encoder to produce an initial recognition hypothesis, which is used for hotword customization (Section~\ref{subsec:hotword}) and as a coarse alignment signal for long-audio context injection (Section~\ref{subsec:long-audio-context}); and (4) an LLM-based decoder that generates the final output conditioned on the audio representations and the CTC hypothesis.

The principal architectural change in this generation is the LLM-based decoder. We replace the previous dense decoder with the Qwen MoE backbone, increasing the linguistic and world-knowledge capacity available to the decoder through sparse expert routing.
The audio encoder and audio adapter remain unchanged from Fun-ASR, consisting of the SenseVoice audio encoder~\citep{DBLP:journals/corr/abs-2407-04051} and a two-layer Transformer adapter.
The CTC decoder is also retained as a lightweight auxiliary module on top of the audio encoder. Rather than performing final autoregressive decoding, it provides an initial recognition hypothesis for hotword customization and a coarse alignment signal for long-audio context injection.

\section{Training}
\label{sec:training}

The training of Qwen-Audio-3.0-ASR comprises five stages:
audio encoder pretraining, audio-language model pretraining, LLM cooldown adaptation, supervised fine-tuning, and reinforcement learning. These stages progressively align acoustic representations with linguistic knowledge and instruction-following capabilities, while further optimizing the model for practical ASR objectives.

\subsection{Pretraining of Audio Encoder}
\label{subsec:pretraining}
\begin{figure}[h]
    \centering
    \includegraphics[width=1\linewidth]{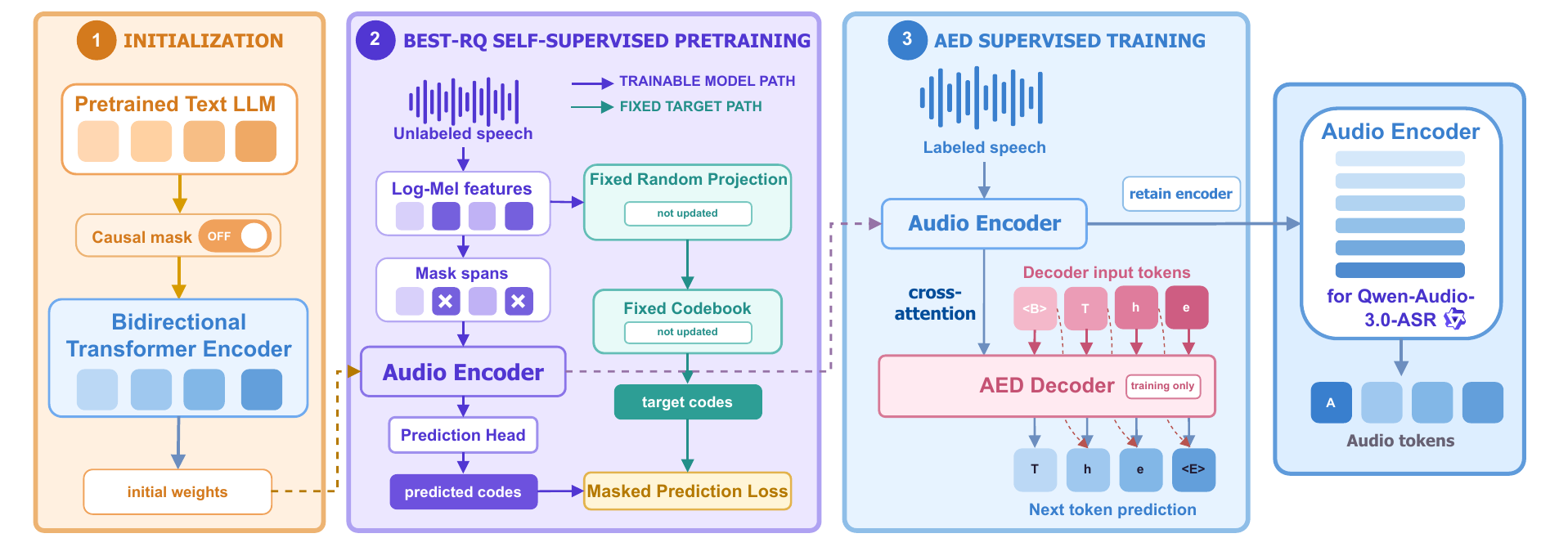}
    \caption{The pretraining pipeline for the audio encoder.}
    \label{fig:encoder}
\end{figure}

As illustrated in Figure~\ref{fig:encoder}, the audio encoder is trained in two consecutive stages: BEST-RQ self-supervised pretraining on large-scale unlabeled speech, followed by supervised training with an attention-based encoder--decoder (AED) objective on labeled audio--text data. Across the two stages, the pretraining corpus comprises tens of millions of hours of audio collected from diverse real-world scenarios.

\paragraph{BEST-RQ self-supervised pretraining.}
We first train the encoder using BEST-RQ~\citep{DBLP:conf/icml/ChiuQZYW22} on large-scale unlabeled speech collected from diverse real-world scenarios, covering domains such as artificial intelligence, biotechnology, e-commerce, education, entertainment, finance, and mobility. Acoustic frames are randomly masked, and the encoder predicts discrete targets generated from the corresponding unmasked features by a fixed random projection and a frozen codebook. This objective enables the encoder to learn contextual speech representations from large-scale and diverse unlabeled audio.
Following our previous study on cross-modal initialization
~\citep{DBLP:journals/corr/abs-2409-17750}, we initialize compatible Transformer layers of the encoder from a pretrained Qwen3 LLM~\citep{DBLP:journals/corr/abs-2505-09388}. Causal attention is replaced with bidirectional attention for speech encoding, while the acoustic front end is initialized separately. The resulting network is subsequently adapted to speech through BEST-RQ pretraining.

\paragraph{Supervised AED training.}
The pretrained encoder is then integrated into an AED model and optimized on large-scale labeled audio--text data, primarily covering Chinese and English. The labeled data are processed through a comprehensive pipeline that includes voice activity detection (VAD), pseudo-label generation by multiple ASR systems, including Paraformer-V2~\citep{DBLP:journals/corr/abs-2409-17746}, Whisper, and SenseVoice~\citep{DBLP:journals/corr/abs-2407-04051}, followed by inverse text normalization (ITN).
Following the general training paradigm of SenseVoice-Large~\citep{DBLP:journals/corr/abs-2407-04051}, an autoregressive decoder attends to the encoder representations and predicts the reference transcription token by token, providing direct acoustic--text supervision. After this stage, the AED decoder is discarded, and the trained encoder is used to initialize the audio encoder of Qwen-Audio-3.0-ASR for subsequent Audio LLM pretraining.

\subsection{Pretraining of Audio LLM}
\label{subsec:pretraining-llm}
\paragraph{Initialization.}
The Audio LLM decoder is initialized from the language-model component of Qwen, while its vision-specific components are not used. The backbone adopts a Mixture-of-Experts (MoE) architecture. This provides substantial linguistic capacity and multilingual knowledge while maintaining relatively low per-token computation, making it suitable for Qwen-Audio-3.0-ASR, including its latency-sensitive streaming variant.

\paragraph{Training data and objective.}
During Audio LLM pretraining, we train the model on approximately 20 million hours of multilingual data constructed through our large-scale data pipeline, consisting primarily of ASR data and interleaved text--audio data. The pretrained audio encoder is kept frozen, while the audio adapter and the LLM decoder are jointly optimized.
The model is trained with a cross-entropy loss over the target text tokens, enabling the audio adapter and LLM decoder to align acoustic representations with the LLM's textual representation space. This stage establishes the
fundamental audio--text alignment required for subsequent ASR adaptation.

The resulting Audio LLM combines the pretrained linguistic knowledge of Qwen with the audio--text alignment learned in this stage. It provides several useful priors for downstream ASR, including multilingual knowledge, world knowledge for resolving semantically ambiguous entities and homophones, and instruction-following capability for handling flexible textual conditions. Meanwhile, the MoE architecture preserves a large total model capacity while activating only part of parameters per token, maintaining relatively low decoding cost.

\subsection{LLM Cooldown Adaptation}
\label{subsec:cooldown}

Following Audio LLM pretraining, we introduce a cooldown stage to further adapt the decoder using a smaller but higher-quality dataset. This stage uses approximately 1 million hours of training data, with the language and dialect distribution rebalanced to improve coverage across different linguistic conditions. In addition to ASR data, the training mixture is enriched with audio understanding, speaker diarization, and SpeechQA data to expose the model to a broader range of tasks and instructions.

To improve data quality, we further filter the ASR portion using VibeVoice-based re-recognition and remove samples whose error rate exceeds 0.3. During this stage, the pretrained SenseVoice audio encoder remains frozen, while the audio adapter and the LLM decoder are jointly optimized using a cross-entropy loss over the target text tokens. Freezing the audio encoder preserves the acoustic representations learned during encoder pretraining, while updating the adapter and decoder allows the model to further refine audio--language alignment and adapt to the higher-quality, better-balanced training mixture.
Compared with Audio LLM pretraining, this cooldown stage emphasizes higher-quality and better-balanced data while introducing more diverse audio tasks and instruction formats. It therefore refines the model's multilingual and instruction-following capabilities before subsequent supervised fine-tuning.

\subsection{Supervised Fine-tuning (SFT)}
\label{subsec:SFT}
SFT integrates the pretrained audio encoder, audio adapter, and the adapted Audio LLM obtained after the preceding Audio LLM pretraining and cooldown stages into a unified, instruction-controlled ASR system. During SFT, we jointly update the audio encoder and audio adapter, while keeping the base LLM parameters frozen and adapting the decoder through Low-Rank Adaptation (LoRA).
The SFT stage uses tens of millions of hours of speech spanning 30 languages and 16 Chinese dialectal varieties. The training mixture combines human-transcribed speech, high-confidence pseudo-labeled data, noise-augmented speech, TTS-generated speech, simulated streaming data, far-field and room-simulated speech, code-switching data, multilingual and dialectal data, and speech from representative business scenarios.
Training examples follow a unified conversational format in which the speech input is accompanied by an instruction containing optional specifications for target languages, polishing, historical transcript context, and hierarchical hotword candidates. These acoustic and textual conditions are jointly processed by the decoder, which is trained with an autoregressive next-token prediction objective. Training sequences are packed to a maximum length of 8,192 tokens. Multiple controls can be combined within a single instruction, enabling configurable recognition behaviors without separate task-specific decoders.

\paragraph{Specified-language transcription.}
To support flexible language control, the SFT mixture includes multilingual, dialectal, and code-switching speech covering the target languages and Chinese dialectal varieties. Each example may be accompanied by a list of target languages that specifies the expected transcription behavior. For monolingual transcription, the instruction requires the model to emit only speech belonging to the specified language and to avoid translating or rewriting speech from other languages into it. For multilingual speech, the model is instructed to preserve each supported language in its original written form. Training on diverse language combinations and code-switching examples enables the model to follow runtime language specifications and reduces unintended cross-language substitutions, particularly for lower-resource languages.

\paragraph{Domain-entity augmentation.} To improve recognition of domain-specific entities that are individually rare in generic ASR data, we augment the SFT mixture with dedicated entity-focused examples. High-value entities are first mined from domains such as artificial
intelligence, finance, healthcare, and IT, and an LLM generates diverse short utterances containing these entities based on retrieved factual context. The generated texts are converted into speech using CosyVoice3~\citep{DBLP:journals/corr/abs-2505-17589}, supplemented by human recordings for terms that are difficult to synthesize reliably. The resulting audio is then re-decoded by the current ASR model, and samples with unambiguous entity-level mismatches are filtered out. These curated examples are incorporated into SFT to increase the model's exposure to rare and rapidly evolving domain terminology.

\paragraph{Native transcription polishing.}
A binary polishing instruction determines whether the target preserves the verbatim spoken form or produces a cleaned transcript. Polishing-enabled examples teach the model to remove non-semantic fillers, collapse stuttering and unnecessary repetitions, resolve explicit self-corrections, and normalize punctuation and textual formatting while preserving the speaker's final intent. The polishing data are constructed from both real and synthetic examples. For real speech, we retain the original audio from existing ASR records and regenerate the textual supervision using a text LLM with production polishing instructions, followed by fidelity filtering to remove samples with semantic loss or unsupported insertions. For synthetic augmentation, an LLM injects fillers, repetitions, and self-corrections into canonical transcripts, and the resulting texts are converted into speech using TTS, followed by ASR-based pronunciation filtering and noise augmentation.
Since the polished transcript itself serves as the ASR target, recognition and polishing are learned within the same autoregressive decoding process, without requiring a cascaded ASR--LLM rewriting pipeline.

\paragraph{Historical-context conditioning.}
For contextual training, transcripts or interaction records from preceding turns are prepended as reference information. The instruction explicitly indicates that the historical context may be incomplete, irrelevant, or incorrect, and that the current audio should remain the primary source of evidence.
The context data are constructed from both real conversations and targeted synthetic examples. For real data, we match ASR records to their original sessions using audio identifiers, restore the chronological order of utterances, and use only preceding turns as context without LLM rewriting.
For synthetic augmentation, we construct examples involving homophone-ambiguous entities together with correct, conflicting, noisy, or empty historical contexts and hotword conditions.
This mixture encourages the model to exploit recurring entities and topic cues when they are informative, while avoiding unconditional copying from the provided context.

\paragraph{Hierarchical hotword conditioning.}
To train hierarchical hotword conditioning, each utterance is paired with two tiers of candidate hotwords. \textbf{P0} contains high-confidence, high-priority terms, including key named entities and other contextually critical expressions, whereas \textbf{P1} contains a broader candidate pool. An LLM assigns candidates to the two tiers by semantically matching them against the reference transcription, promoting unambiguous contextual matches
to P0 and retaining the remaining candidates in P1.
During SFT, both candidate sets are included in the instruction as auxiliary conditioning inputs. The model learns to place greater preference on P0 candidates when supported by the acoustic and contextual evidence, while
using P1 as a broader source of possible entities. This design improves long-tail hotword recall without forcing acoustically unsupported candidates into the transcript.

Together, these data construction strategies and instruction-conditioned training objectives enable Qwen-Audio-3.0-ASR to jointly model acoustic evidence and user-specified controls within a unified framework. At inference time, language selection, native polishing, historical context, and P0/P1 hotwords can be enabled independently or composed within the same request, as illustrated in Figure~\ref{fig:model}.

\subsection{Reinforcement Learning (RL)}
\label{subsec:RL}

Following SFT, we apply reinforcement learning to further optimize Qwen-Audio-3.0-ASR for practical ASR objectives that are difficult to capture with maximum-likelihood training alone.

\paragraph{RL training data.}
We construct a compact, high-quality RL dataset focusing on challenging real-world cases. The dataset includes: (1) hard cases identified from disagreements between the base model and multiple external ASR systems, including Whisper, FireRed-ASR, and SenseVoice; (2) long-duration utterances exceeding 20 seconds; (3) hallucination-related samples, including both actual hallucinations and genuine long repetitions that may superficially resemble hallucinations; (4) keyword- and hotword-focused samples; and (5) regular ASR data included to mitigate catastrophic forgetting.
Different sample types are associated with task-specific reward functions during RL, as described below.

\paragraph{The FunVerl-ASR Framework.}
We develop \textbf{FunVerl-ASR}, a fully asynchronous reinforcement learning framework purpose-built for large audio-language models (LALMs). Existing RL frameworks for LLMs, such as verl~\citep{sheng2024hybridflow} and TRL~\citep{vonwerra2022trl}, assume text-only inputs and do not natively support the audio encoder that Qwen-Audio-3.0-ASR requires to convert speech waveforms into token-aligned embeddings. FunVerl-ASR addresses this gap by extending the verl infrastructure with audio-aware data pipelines, a dedicated audio feature extraction stage, and an asynchronous training loop that decouples rollout generation from policy optimization.

\begin{figure}[htbp]
    \centering
    \includegraphics[width=1.0\linewidth]{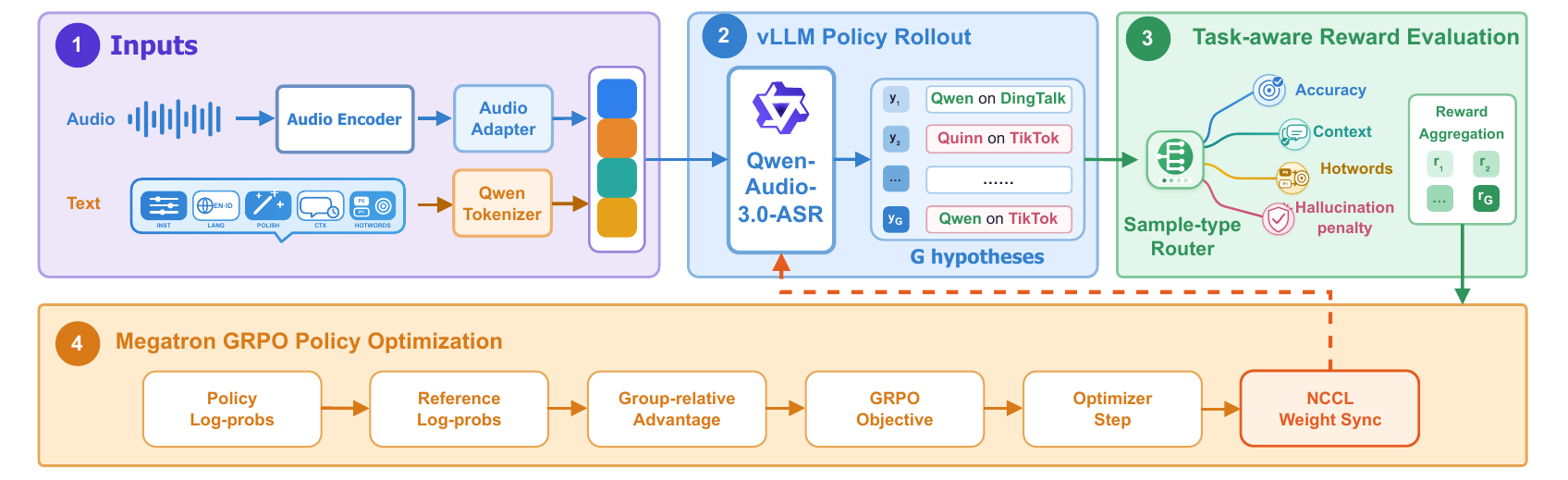}
    \caption{Overview of the FunVerl-ASR fully asynchronous RL framework for Qwen-Audio-3.0-ASR. The audio processor extracts Mel features and tokenizes instruction text; the vLLM rollout engine runs the frozen audio encoder internally and generates hypotheses from merged audio-text embeddings; the reward module scores each hypothesis; and the Megatron policy trainer performs GRPO updates with NCCL-based weight synchronization back to the rollout engine.}
    \label{fig:funverl}
\end{figure}

As illustrated in Figure~\ref{fig:funverl}, FunVerl-ASR decomposes RL training into four concurrent modules orchestrated by Ray:

\begin{itemize}[leftmargin=*,noitemsep]
\item \textbf{Audio Processor.} Raw audio waveforms are converted into Mel-spectrogram features by the audio processor, while instruction text (including context, hotwords, and other flexible prompts) is tokenized by the Qwen tokenizer. Both outputs are passed together to the rollout engine.

\item \textbf{vLLM Rollout Engine.} The vLLM-based inference engine receives Mel features and token IDs jointly. Inside the engine, the frozen audio encoder and adapter transform Mel features into frame-level audio embeddings, which are then merged into the text embedding sequence. The combined embedding sequence is decoded by the MoE LLM to generate $G$ candidate hypotheses per prompt. 

\item \textbf{Reward Module.} Each generated hypothesis is evaluated by a configurable reward module that supports rule-based scoring (CER, keyword recall, hotword hit, context entity match, hallucination detection, language consistency), LLM-judge scoring, and per-sample reward routing for mixed-domain batches.

\item \textbf{Megatron Policy Trainer.} The actor model is trained using Megatron-Core with combined tensor, pipeline, and expert parallelism. After each gradient update, an NCCL-based checkpoint engine synchronizes the updated LLM decoder weights to the vLLM rollout workers, maintaining near-on-policy training without requiring full model reloading. The audio encoder remains frozen throughout RL and is not included in the parameter synchronization.
\end{itemize}

The key architectural innovation of FunVerl-ASR is its fully asynchronous training loop: the rollout engine continuously generates trajectories and pushes them into a message queue, while the policy trainer consumes batches from the queue and performs gradient updates independently. This eliminates the synchronous barrier between generation and training that dominates wall-clock time in conventional RL frameworks. A staleness threshold ensures that trajectories generated under excessively outdated policies are discarded, bounding the off-policy gap. The default configuration generates 8 hypotheses per prompt with a mini-batch size of 32 prompts per actor update.

We evaluate FunVerl-ASR's training efficiency on a 4-node cluster (32 A100 GPUs, split equally between rollout and training). The entire RL training of Qwen-Audio-3.0-ASR completes within one day using this configuration.

\paragraph{GRPO-based RL for ASR.}
Based on the FunVerl-ASR framework, we employ Group Relative Policy Optimization (GRPO)~\citep{shao2024deepseekmath} as the RL algorithm.
GRPO is a lightweight policy-based method that generates a group of responses $\{o_i\}_{i=1}^{G}$ per prompt and computes rule-based rewards $\{R_i\}_{i=1}^G$ without requiring a separately trained value network. The group-level rewards are normalized to compute the advantage:

\begin{equation}
\hat{A}_{i,t}=\frac{R_i-\text{mean}\!\left(\{R_j\}_{j=1}^G\right)}{\text{std}\!\left(\{R_j\}_{j=1}^G\right)}
\end{equation}
The policy is optimized with a clipped objective and a KL penalty term:

\begin{equation}
\begin{split}
L_{\mathrm{GRPO}}(\theta) = 
\frac{1}{G}\sum_{i=1}^G \frac{1}{|o_i|}\sum_{t=1}^{|o_i|} 
&\min\!\bigl(r_{i,t}(\theta)\,\hat{A}_{i,t},\;
\operatorname{clip}\bigl(r_{i,t}(\theta),1-\varepsilon,1+\varepsilon\bigr)\,\hat{A}_{i,t}\bigr) \\
& - \beta\,D_{\mathrm{KL}}\!\bigl(\pi_\theta\|\pi_{\mathrm{ref}}\bigr)
\end{split}
\end{equation}
where $r_{i,t}(\theta)={\pi_{\theta}(o_{i,t}\mid q,\,o_{i,<t})}/{\pi_{\theta_{\mathrm{old}}}(o_{i,t}\mid q,\,o_{i,<t})}$.

When WER is used as the reward, GRPO is reduced to a form closely related to Minimum Word Error Rate (MWER) training, a well-established criterion in the ASR community. Beyond basic WER, we design a composite reward function with per-sample routing to handle mixed-domain training batches:

\begin{itemize}[leftmargin=*,noitemsep]
    \item \textbf{ASR Accuracy ($R_i^1$).} $1 - \text{WER}(y^*, y)$ as the primary reward, with range $[0, 1]$.
    
    \item \textbf{Keyword Accuracy and Recall ($R_i^2$).} Keyword recall encourages correct entity recognition; keyword accuracy penalizes insertion errors to balance precision and recall. Keywords are either manually annotated or LLM-identified.
    
    \item \textbf{Hotword Reward ($R_i^3$).} For hotword-conditioned samples, a dedicated reward scores recognition of P0/P1 hotwords provided in the prompt, with asymmetric positive/negative gamma weighting.
    
    \item \textbf{Context Reward ($R_i^4$).} For context-conditioned samples, rewards hit/miss of context-derived entities with configurable penalties for false insertions.
    
    \item \textbf{Hallucination Suppression ($R_i^5$).} Hallucinated content is detected via pattern matching and penalized proportionally to the hallucinated span length.
    
\end{itemize}

FunVerl-ASR's reward routing mechanism selects the appropriate reward configuration per sample based on its data source (general, dialect, hotword, or context), enabling a single training run to jointly optimize across all production scenarios. 

\section{Deployment}
\label{sec:deployment}
\subsection{Multilingual and Dialectal ASR}
\label{subsec:multilinguality}
The availability of training data varies widely across different languages. Resource-rich languages such as Chinese and English have abundant data, whereas languages such as Vietnamese and Thai have comparatively limited resources. Previous releases addressed this by training a separate multilingual model, \textbf{Fun-ASR-ML}, alongside the primary Chinese-English model; the latest such model supported 31 languages, which is deployed as a checkpoint independent from the Chinese-English model.

Qwen-Audio-3.0-ASR replaces this per-market-model approach with a single \textbf{flexible-instruction} interface: a target-language list is passed as part of the input instruction, and the model adapts its decoding to the specified language combination at inference time, rather than requiring a checkpoint switch. Language combinations anchor on Chinese, English, or Japanese as the primary language and freely add secondary languages -- for example ``Chinese+English+Japanese'', ``Chinese+French'', or ``English+Indonesian'' -- covering cross-border customer service, multilingual meetings, and similar scenarios. Through this single interface, Qwen-Audio-3.0-ASR supports 30
languages across multiple regions: East Asian languages (Chinese, Japanese, and Korean); Southeast Asian languages (Vietnamese, Thai, Indonesian, Malay, and Tagalog); Hindi and Arabic; and European languages (English, French, German, Spanish, Portuguese, Russian, Italian, Dutch, Swedish, Danish, Finnish, Greek, Polish, Czech, Hungarian, Romanian, Bulgarian, Croatian, Slovak, and Norwegian).
The same instruction interface additionally supports transcription and Mandarin translation for 16 Chinese dialectal varieties spanning eight major dialect regions.

\subsection{Low-Latency Streaming Recognition}
\label{subsec:streaming recognition}
To support interactive speech applications, we introduce a hybrid streaming and full-context recognition framework. The streaming model processes incoming audio incrementally and continuously updates a provisional transcript for immediate on-screen display. Once the utterance is complete, the full-context large model re-evaluates the entire acoustic sequence and refreshes the provisional output with a more accurate finalized transcript. This coarse-to-fine workflow combines immediate user feedback with the recognition quality enabled by complete acoustic context.

Low-latency recognition is achieved by processing audio in short chunks with limited right context, allowing text to appear before an utterance is complete. At inference time, the chunk size and right-context length can be configured independently, providing multiple operating points without changing the overall recognition framework. Smaller contexts prioritize responsiveness, whereas larger contexts provide more acoustic evidence and improve recognition accuracy. The resulting system can therefore adapt to the latency and accuracy requirements of different application scenarios.

\subsection{Message ASR for Context-Aware Transcription}
\label{subsec:message asr}

\begin{figure*}[t]
    \centering
    \includegraphics[width=\textwidth]
    {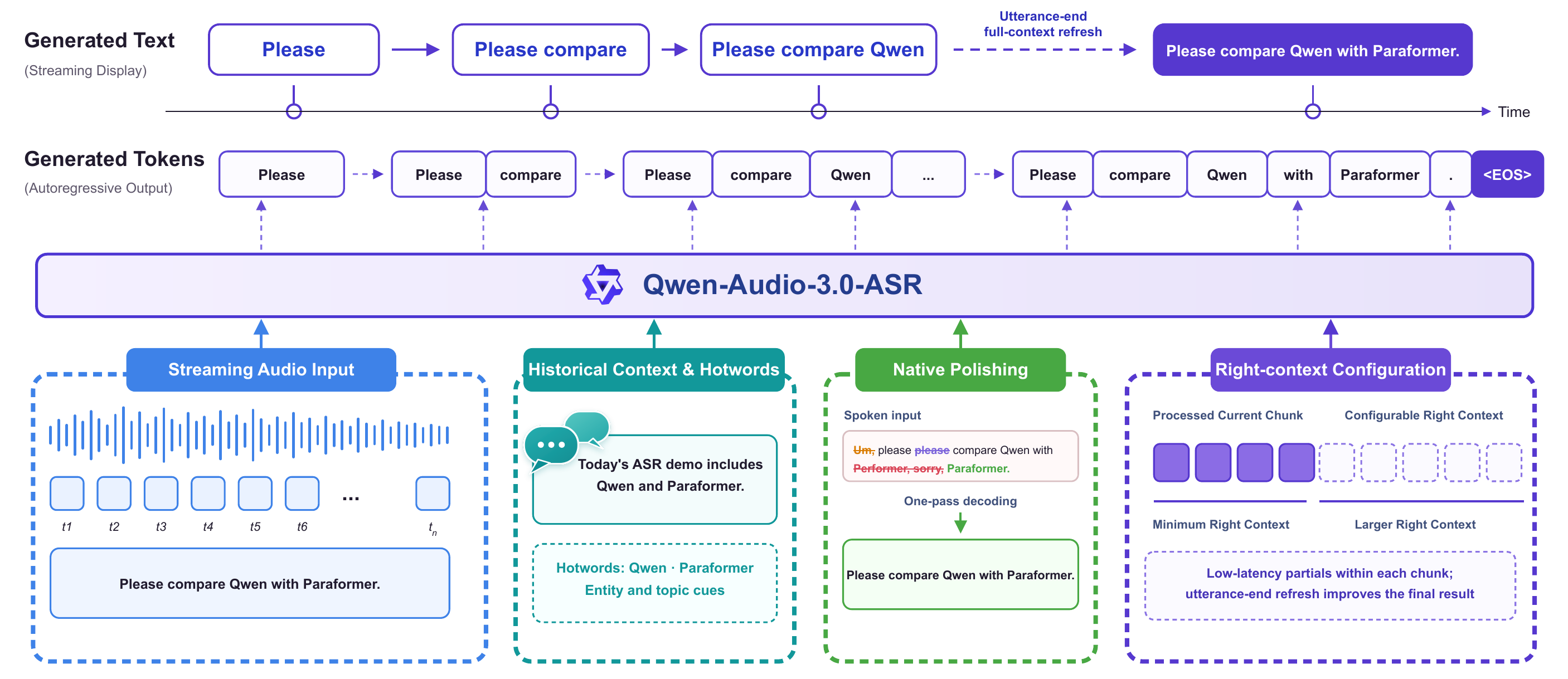}
    \caption{
        Overview of the Message ASR workflow built on
        Qwen-Audio-3.0-ASR. Streaming audio is decoded incrementally to
        provide responsive partial transcripts. Historical context
        supplies entity and topic cues, native polishing improves
        readability within the same decoding pass, and configurable
        right context balances streaming latency and recognition
        accuracy. At utterance end, the model refreshes the displayed
        hypothesis using the complete available context.
    }
    \label{fig:message-asr-overview}
\end{figure*}

Message ASR is a deployment profile of Qwen-Audio-3.0-ASR designed for real-world transcription scenarios, including voice input, meetings, interviews, and customer-service conversations. It turns spontaneous or extended speech into readable text that can be reviewed, shared, or archived with minimal editing. Rather than introducing a separate recognition stack, it composes the model's instruction-controlled capabilities into one decoding workflow. 
As illustrated in Figure~\ref{fig:message-asr-overview}, streaming audio chunks, historical context, native polishing instructions, and configurable right context are jointly consumed by Qwen-Audio-3.0-ASR. The model emits low-latency partial hypotheses during streaming and performs a full-context refresh after utterance-end detection, producing a more accurate and readable final transcript without introducing a separate recognition or rewriting pipeline.
Incoming audio is processed incrementally to maintain a responsive provisional transcript; after endpoint detection, the complete utterance or speech segment is re-evaluated and the on-screen text is refreshed. Historical conversation turns or preceding transcript segments provide topic and entity continuity, while P0/P1 hotwords distinguish a small set of high-confidence terms from a broader candidate pool. These signals guide the decoder only when they agree with acoustic and semantic evidence, reducing substitutions of names, products, and technical terms without blindly copying the prompt.

The same interface supports 30 languages and 16 dialectal varieties across eight major Chinese dialect regions. A target-language list can be supplied at runtime, allowing applications to preserve multilingual speech in its original writing system and support cross-lingual conversations without switching checkpoints. Native polishing converts disfluent speech into a readable transcript within the recognition pass by removing fillers and redundant repetitions, resolving explicit self-corrections, and normalizing punctuation while retaining the speaker's final intent. Because all conditions are consumed by the same model, applications can change language, context, hotword priority, and polishing behavior per request without maintaining parallel pipelines or adding a post-recognition rewrite service. These controls can be enabled independently or combined for different product modes, such as verbatim transcription, polished dictation, contextual meeting transcription, interview recording, customer-service documentation, and low-latency streaming. Message ASR therefore serves as a context-aware transcription interface that balances immediate visual feedback, long-tail entity accuracy, and final-text readability. Audio remains the primary source of evidence throughout the pipeline, which is essential when conversation history or user-provided hotwords are incomplete, outdated, or incorrect.

\section{Evaluations}
\label{sec:exp}

We evaluate Qwen-Audio-3.0-ASR on both open-source ASR benchmark datasets and real-world industry evaluation sets. For the open-source evaluation, we use corresponding test sets of AISHELL-1~\citep{bu2017aishell1opensourcemandarinspeech}, AISHELL-2~\citep{du2018aishell2transformingmandarinasr},  FLEURS~\citep{conneau2022fleursfewshotlearningevaluation}, LibriSpeech~\citep{7178964}, WeNetSpeech~\citep{zhang2022wenetspeech10000hoursmultidomain}, GigaSpeechBench~\citep{tu2026gigaspeechbench} and CommonVoice~\citep{ardila2020commonvoice} data sets.
In addition to these public benchmarks, we evaluate the system on a broad set of internal industrial test sets designed to reflect practical deployment conditions. These evaluations cover industry-specific entities and long-tail terminology, hotword customization, long-audio contextual recognition, transcription polishing, and low-latency streaming recognition. The internal test sets span diverse acoustic conditions, speaking styles, application domains, and usage scenarios, providing a complementary assessment of the system's production-oriented capabilities.

\subsection{Public Benchmark Results}

\subsubsection{Chinese and English ASR Benchmarks}

We evaluate Qwen-Audio-3.0-ASR against a range of publicly released models and commercial ASR systems on widely used Chinese and English benchmarks, as shown in Table~\ref{tab:asr-os}. Panel A compares our model with results published in official model repositories or technical reports, whereas Panel B compares systems evaluated using our unified API-based evaluation pipeline. We therefore discuss the two panels separately.

\paragraph{Comparison with officially reported results.}
As shown in Panel A, Qwen-Audio-3.0-ASR has the lowest reported error rate on four of the seven test sets. Compared with the strongest officially reported baseline on each of these test sets, its error rate is lower by 0.08 percentage points (3.8\% relative) on AISHELL-2, 0.60 points (22.4\% relative) on FLEURS-zh, 0.18 points (7.4\% relative) on LibriSpeech-other, and 0.36 points (7.7\% relative) on WeNetSpeech-net. It also remains within 0.02 percentage points of the best officially reported result on LibriSpeech-clean. Because these baseline results are collected from different public sources, they may reflect source-specific preprocessing, reference transcripts, and scoring conventions. These numerical differences should therefore be interpreted with caution, particularly for WeNetSpeech-net, where our evaluations use corrected reference transcripts.

\paragraph{Comparison under our unified evaluation pipeline.}
As shown in Panel B, Qwen-Audio-3.0-ASR achieves the lowest error rate on four of the seven test sets. Relative to the strongest competing system evaluated using the same pipeline, it reduces the error rate by 0.30 percentage points (12.6\% relative) on FLEURS-zh, 0.78 points (15.8\% relative) on FLEURS-en, 0.16 points (11.9\% relative) on LibriSpeech-clean, and 0.27 points (10.8\% relative) on LibriSpeech-other. It additionally remains within 0.09 percentage points of the best result on AISHELL-2. These results demonstrate competitive recognition performance across both Chinese and English benchmarks.

\begin{table*}[t]
\centering
\caption{
Evaluation results on public ASR benchmark test sets.
We report CER (\%) for Chinese test sets and WER (\%) for English
test sets. For the competing systems, Panel A presents results taken
from official public repositories or technical reports, whereas
Panel B presents results obtained using our unified API-based
evaluation pipeline on the same test sets. Qwen-Audio-3.0-ASR is
repeated in both panels for reference. Lower values indicate better
performance, and the best result within each panel on each test set
is shown in bold. Because the results in Panel A may involve
source-specific preprocessing and scoring conventions, direct
quantitative comparisons are primarily made within Panel B. For WeNetSpeech-net, all results in Panel B and our model's results in Panel A are evaluated using corrected reference transcripts, following the audio--label mismatch discussion in \href{https://github.com/wenet-e2e/WenetSpeech/issues/63} {WenetSpeech issue~\#63}. The competing-system results in Panel A are retained as reported in their original sources.
}
\label{tab:asr-os}

\small
\setlength{\tabcolsep}{3.5pt}
\renewcommand{\arraystretch}{1.13}

\begin{tabular*}{\textwidth}{
    @{\extracolsep{\fill}}
    llcccccc
    @{}
}
\toprule

\rowcolor{QwenPurpleLight}
\multicolumn{8}{c}{
    \color{QwenPurple}\bfseries
    Panel A: Officially Reported Results
} \\

\midrule

Test set
& Metric
& \shortstack{Whisper\\-large-v3}
& \shortstack{Kimi\\-Audio}
& \shortstack{Step\\-Audio 2}
& \shortstack{FireRed\\-ASR}
& \shortstack{LongCat-Flash-\\Omni Instruct}
& \shortstack{Qwen-Audio\\-3.0-ASR} \\

\midrule

AISHELL-1
& CER
& 4.72
& 0.71
& 0.63
& \textbf{0.54}
& 0.63
& 1.03 \\

AISHELL-2
& CER
& 4.68
& 2.86
& 2.10
& 2.58
& 2.78
& \textbf{2.02} \\

FLEURS-zh
& CER
& 5.18
& 3.11
& 2.68
& 4.81
& 3.99
& \textbf{2.08} \\

FLEURS-en
& WER
& 6.23
& 6.99
& \textbf{3.03}
& 10.79
& 5.02
& 4.17 \\

LibriSpeech-clean
& WER
& 1.86
& 1.32
& \textbf{1.17}
& 1.84
& 1.57
& 1.19 \\

LibriSpeech-other
& WER
& 3.43
& 2.63
& 2.42
& 4.52
& 4.01
& \textbf{2.24} \\

WeNetSpeech-net
& CER
& 11.89
& 6.45
& 4.67
& 4.94
& 6.09
& \textbf{4.31} \\

\bottomrule
\end{tabular*}

\vspace{8pt}

\begin{tabular*}{\textwidth}{
    @{\extracolsep{\fill}}
    llcccccc
    @{}
}
\toprule

\rowcolor{QwenPurpleLight}
\multicolumn{8}{c}{
    \color{QwenPurple}\bfseries
    Panel B: Results from Our API Evaluation
} \\

\midrule

Test set
& Metric
& \shortstack{GPT-4o\\Transcribe}
& Azure
& Doubao-ASR
& \shortstack{Fun-ASR\\-Flash}
& \shortstack{Tencent\\Hy-ASR-3.0-\\preview}
& \shortstack{Qwen-Audio\\-3.0-ASR} \\

\midrule

AISHELL-1
& CER
& 3.34
& 1.60
& 1.54
& \textbf{0.76}
& 0.78
& 1.03 \\

AISHELL-2
& CER
& 4.47
& 3.24
& 2.66
& \textbf{1.93}
& 2.15
& 2.02 \\

FLEURS-zh
& CER
& 6.14
& 5.51
& 8.18
& 5.52
& 2.38
& \textbf{2.08} \\

FLEURS-en
& WER
& 5.79
& 5.47
& 6.50
& 4.95
& 5.45
& \textbf{4.17} \\

LibriSpeech-clean
& WER
& 1.58
& 1.90
& 3.36
& 1.35
& 1.60
& \textbf{1.19} \\

LibriSpeech-other
& WER
& 3.87
& 3.54
& 4.46
& 2.51
& 2.68
& \textbf{2.24} \\

WeNetSpeech-net
& CER
& 12.27
& 4.58
& 4.81
& \textbf{3.83}
& 4.16
& 4.31 \\

\bottomrule
\end{tabular*}

\end{table*}

\begin{table*}[t]
\centering
\caption{
Error rates (\%) of multilingual ASR systems on Common Voice 15, FLEURS, and GigaSpeechBench. The evaluation metric for each language is given in the second column. Japanese CER is computed following the normalization and scoring protocol specified in \href{https://github.com/QwenAudio/Fun-ASR/pull/165}
{Fun-ASR PR \#165}. Lower values indicate better performance. Macro averages are calculated only for systems with results on all languages included in the corresponding benchmark. Bold values denote the lowest error rate among systems evaluated on the corresponding language. N/A denotes a language unsupported by the model.}
\label{tab:multilingual}

\scriptsize
\setlength{\tabcolsep}{8pt}
\renewcommand{\arraystretch}{1.13}

\begin{adjustbox}{max width=\textwidth}
\begin{tabular}{cccccccc}
\toprule

Language
& Metric
& \shortstack{Doubao-ASR}
& \shortstack{Tencent\\Cloud ASR}
& \shortstack{Gemini\\3.1 Pro}
& \shortstack{GPT-4o\\Transcribe}
& \shortstack{Azure}
& \shortstack{Qwen-Audio\\-3.0-ASR} \\
\midrule

\rowcolor{QwenPurpleLight}
\multicolumn{8}{c}{
    \color{QwenPurple}\bfseries
    GigaSpeechBench
} \\

Indonesian
& WER
& 27.73 & 32.55 & 24.47 & 39.52 & 25.02
& \textbf{23.68} \\

Japanese
& CER
& 29.47 & 34.56 & 39.12 & 45.46 & 29.94
& \textbf{28.48} \\

Thai
& CER
& 17.65 & 31.40 & 17.37 & 53.55 & 17.08
& \textbf{16.22} \\

Filipino
& WER
& \textbf{25.36} & 33.38 & 37.41 & 37.73 & N/A
& 29.45 \\

Vietnamese
& WER
& \textbf{9.53} & 9.61 & 15.55 & 32.89 & 13.25
& 11.54 \\

Korean
& CER
& 14.52 & 15.00 & 19.70 & 32.52 & 16.39
& \textbf{13.50} \\

Malay
& WER
& 35.05 & 41.06 & 40.19 & 51.59 & N/A
& \textbf{34.65} \\

\cmidrule(lr){1-8}
Macro average
& --
& 22.76 & 28.22 & 27.69 & 41.89 & --
& \textbf{22.50} \\

\midrule

\rowcolor{QwenPurpleLight}
\multicolumn{8}{c}{
    \color{QwenPurple}\bfseries
    Common Voice 15
} \\

Indonesian
& WER
& 4.25 & 7.79 & 12.92 & 8.74 & 4.78
& \textbf{3.66} \\

Japanese
& CER
& 6.44 & 12.02 & 11.52 & 11.22 & 6.81
& \textbf{5.64} \\

Thai
& CER
& 2.32 & 4.17 & 12.33 & 6.64 & 2.39
& \textbf{1.56} \\

Vietnamese
& WER
& 6.31 & \textbf{5.67} & 10.64 & 15.50 & 6.64
& 8.29 \\

Korean
& CER
& 3.89 & 8.16 & 4.86 & 5.80 & 3.32
& \textbf{3.09} \\

Spanish
& WER
& 5.42 & 4.95 & 3.97 & 4.18 & 3.40
& \textbf{3.09} \\

French
& WER
& 8.89 & 22.06 & 10.51 & 9.26 & 7.73
& \textbf{6.69} \\

\cmidrule(lr){1-8}
Macro average
& --
& 5.36 & 9.26 & 9.54 & 8.76 & 5.01
& \textbf{4.57} \\

\midrule

\rowcolor{QwenPurpleLight}
\multicolumn{8}{c}{
    \color{QwenPurple}\bfseries
    FLEURS
} \\

Indonesian
& WER
& 5.24 & 12.77 & 3.72 & 3.73 & 3.12
& \textbf{2.89} \\

Japanese
& CER
& 3.16 & 15.39 & 3.95 & 2.98 & 2.08
& \textbf{1.61} \\

Thai
& CER
& 6.18 & 11.38 & 6.05 & 4.99 & \textbf{4.65}
& 6.73 \\

Filipino
& WER
& 10.34 & 18.05 & 8.33 & \textbf{7.42} & N/A
& 11.48 \\

Vietnamese
& WER
& 6.32 & 15.19 & 4.70 & 3.23 & \textbf{2.79}
& 4.95 \\

Korean
& CER
& 4.86 & 8.93 & 4.24 & 4.06 & \textbf{3.89}
& 4.94 \\

Malay
& WER
& 8.14 & 17.13 & 4.22 & \textbf{4.21} & N/A
& 8.79 \\

Spanish
& WER
& 5.31 & 8.09 & 2.62 & 2.13 & 2.06
& \textbf{1.23} \\

Portuguese
& WER
& 7.19 & 13.71 & 3.46 & 2.80 & 2.59
& \textbf{1.87} \\

\cmidrule(lr){1-8}
Macro average
& --
& 6.30 & 13.40 & 4.59 & \textbf{3.95} & --
& 4.94 \\

\bottomrule
\end{tabular}
\end{adjustbox}

\vspace{3pt}

\end{table*}

\subsubsection{Multilingual ASR Benchmarks}
To further assess the multilingual capabilities of the system, we evaluate Qwen-Audio-3.0-ASR on three public multilingual benchmarks, including GigaSpeechBench, Common Voice 15, and FLEURS, using WER or CER according to the language-specific evaluation protocol. As shown in Table~\ref{tab:multilingual}, GigaSpeechBench covers seven East and Southeast Asian languages, on which Qwen-Audio-3.0-ASR achieves a macro-average error rate of 22.50\%, the lowest among systems evaluated on all seven languages, and records the lowest error rate on five of them. On Common Voice 15, which covers five Asian languages together with Spanish and French, the model achieves the lowest macro-average error rate of 4.57\% and the lowest error rate on six of the seven languages. On FLEURS, which evaluates seven Asian languages together with Spanish and Portuguese, the results are more mixed: Qwen-Audio-3.0-ASR records the lowest error rate on four of the nine languages, with a macro-average error rate of 4.94\%.

\subsection{Chinese Dialect Recognition and Consistency}
\label{subsec:dialect-results}

\begin{figure*}[t]
    \centering
    \includegraphics[width=\textwidth]
    {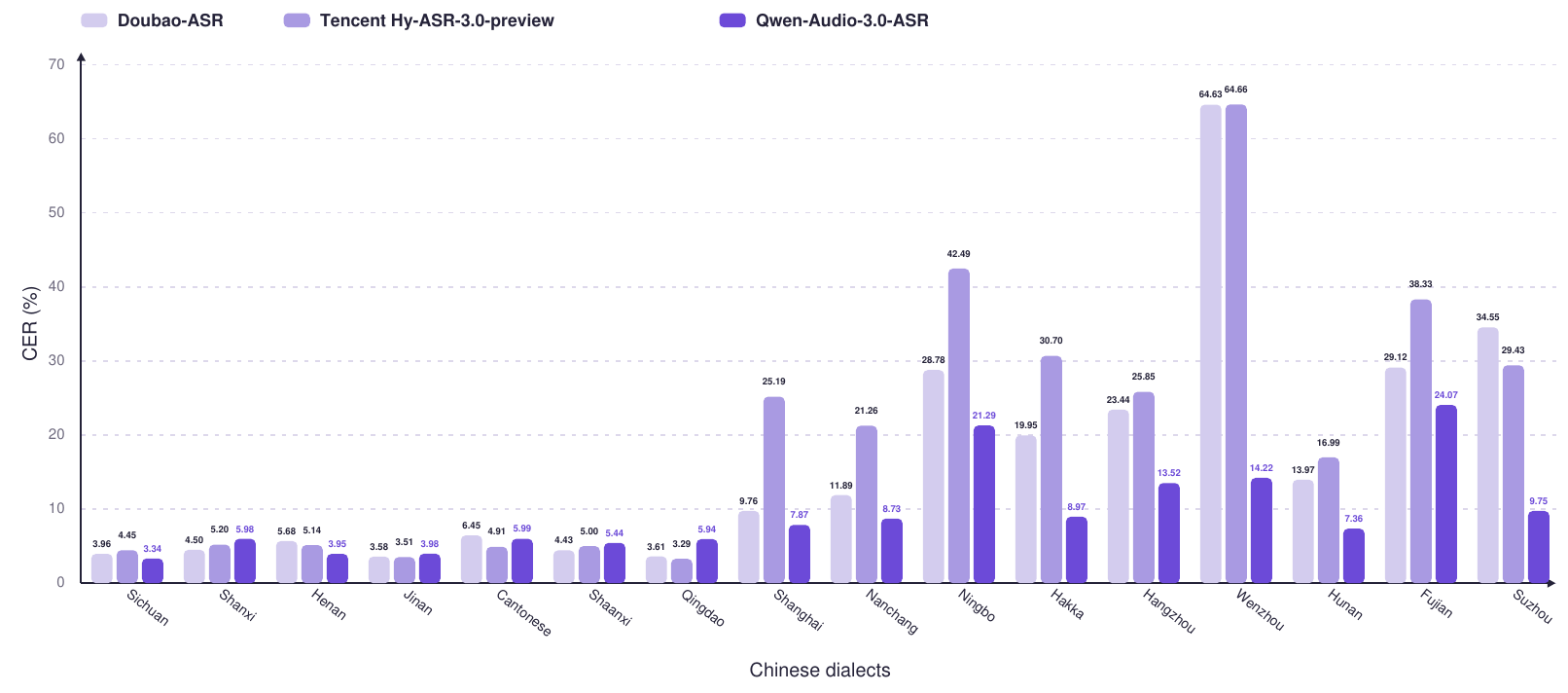}
    \caption{
    Character error rate (CER, \%) on an internal evaluation suite
    covering 16 Chinese dialects under the ASR and AST settings
    described in Section~\ref{subsec:dialect-results}.
    We compare Doubao-ASR, Tencent Hy-ASR-3.0-preview, and
    Qwen-Audio-3.0-ASR. Lower values indicate better performance.
    }
    \label{fig:dialect-cer}
\end{figure*}

\begin{figure*}[t]
    \centering
    \includegraphics[width=\textwidth]
    {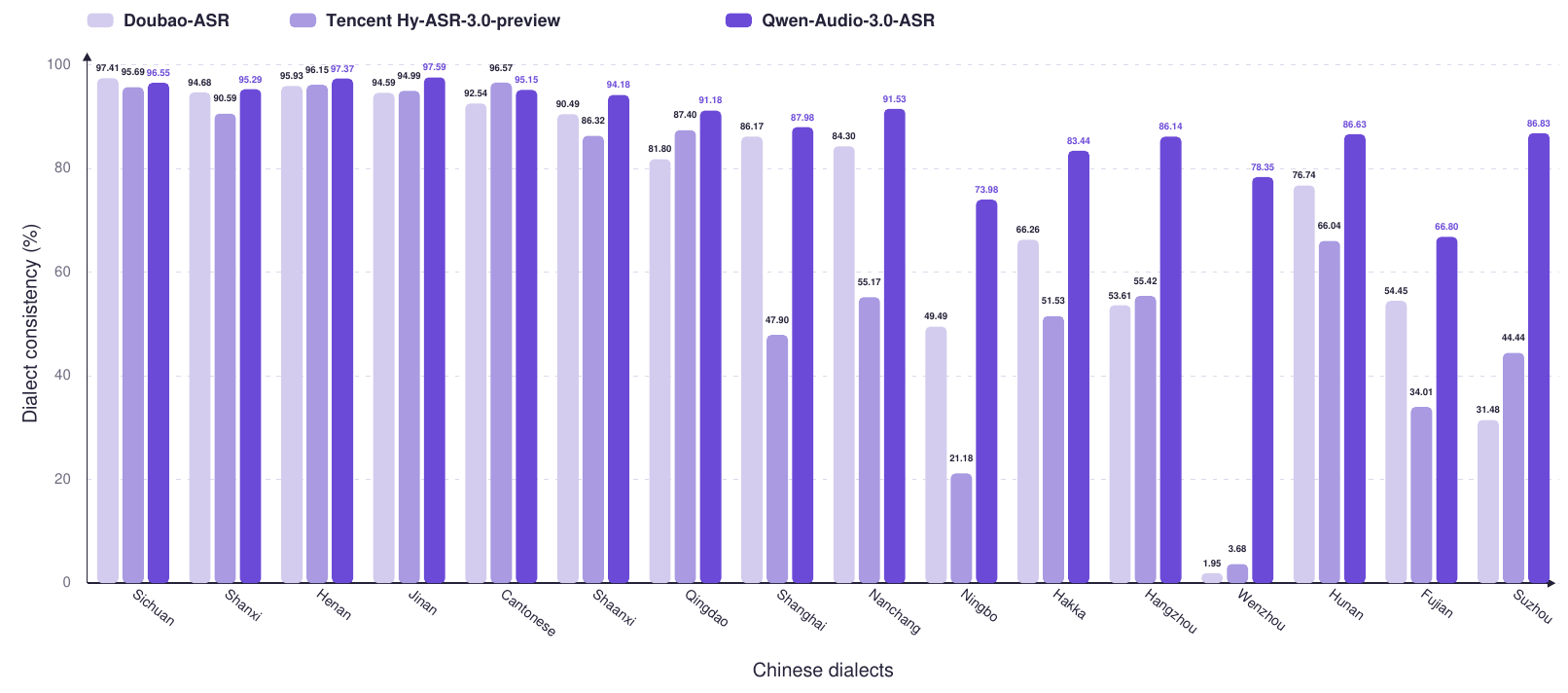}
    \caption{
    Dialect-consistency rates (\%) on an internal evaluation suite
    covering 16 Chinese dialects. Each utterance is scored from 0 to
    10 by Qwen3.7-Max, and an output with a score of at least 6 is
    counted as consistent. We compare Doubao-ASR,
    Tencent Hy-ASR-3.0-preview, and Qwen-Audio-3.0-ASR.
    Higher values indicate better performance.
    }
    \label{fig:dialect-consistency}
\end{figure*}

We evaluate Chinese dialect recognition on an internal evaluation suite covering 16 Chinese dialects. The evaluation considers two complementary dimensions: transcription accuracy, measured by CER, and task consistency, as shown in Figure~\ref{fig:dialect-cer} and Figure~\ref{fig:dialect-consistency}.

The evaluation suite contains two task configurations. Sichuan, Shanxi, Henan, Jinan, Cantonese, Shaanxi, and Qingdao are evaluated using the ASR setting, in which the
model is required to faithfully transcribe the speech in the
corresponding dialect. Shanghai, Nanchang, Ningbo, Hakka, Hangzhou,
Wenzhou, Hunan, Fujian, and Suzhou are evaluated using the automatic speech translation (AST) setting, in which the model is required to translate the dialectal speech into natural Mandarin.

Dialect consistency is evaluated by Qwen3.7-Max using a unified 0--10 scoring rubric. For each utterance, the evaluator first assesses semantic fidelity, including the preservation of the core intent and key information such as actions, entities, numbers, negation, and logical relations, and then checks compliance with the corresponding task requirements. For ASR, the output should faithfully preserve the target-dialect transcription, and inappropriate conversion into Mandarin is penalized when the reference contains explicit dialectal expressions. For AST, the output should be a natural Mandarin translation without unintended dialect residue. An utterance is considered consistent if its score is at least 6. The consistency rate is computed as
\[
\mathrm{Consistency}
=
\frac{100}{N}
\sum_{i=1}^{N}
\mathbf{1}\!\left[s_i \geq 6\right],
\]
where $s_i$ denotes the consistency score of the $i$-th utterance, $N$ is the total number of utterances, and $\mathbf{1}[\cdot]$ is the indicator function.

As shown in Figure~\ref{fig:dialect-cer}, Qwen-Audio-3.0-ASR achieves the lowest CER on 11 of the 16 dialect subsets. Its macro-average CER is 9.40\%, compared with 16.77\% for Doubao-ASR and 20.40\% for Tencent Hy-ASR-3.0-preview, corresponding to absolute reductions of 7.37 and 11.00 percentage points, respectively. The improvements are especially pronounced on challenging dialects. For example, on Wenzhou, Qwen-Audio-3.0-ASR reduces CER to 14.22\%, compared with 64.63\% for Doubao-ASR and 64.66\% for Tencent Hy-ASR-3.0-preview. Substantial improvements are also observed on Shanghai, Nanchang, Ningbo, Hakka, Hangzhou, Hunan, Fujian, and Suzhou.

\begin{figure}[h]
    \centering
    \includegraphics[width=\linewidth]{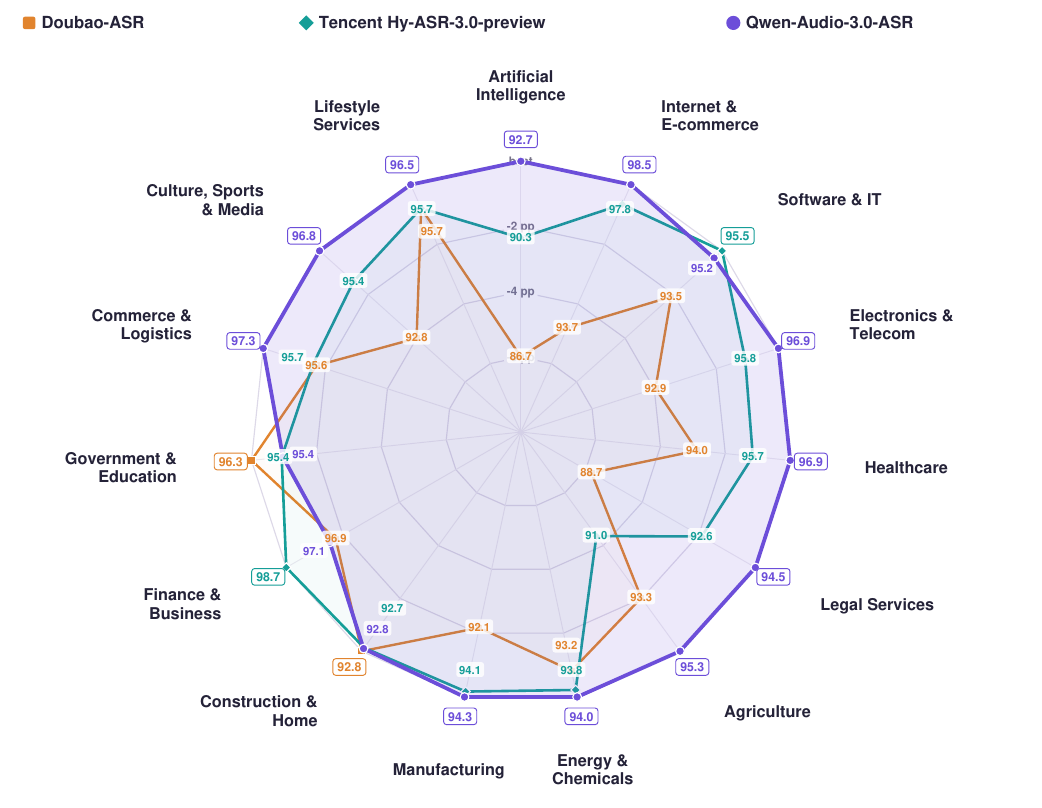}
    \caption{Industry-domain entity recall across 15 domains. For visualization, radial positions are normalized independently within each domain so that the best-performing system lies on the outer ring; numerical labels report the original entity-recall values (\%).}
    \label{fig:entity-recall}
\end{figure}

\subsection{Industry-domain Entity Recognition}
\label{subsec:industry-terms}

Beyond general-purpose recognition accuracy, real-world ASR systems require reliable recognition of domain-specific terminology, including financial terms, medical terminology, software and IT vocabulary, organization names, and other long-tail entities that occur relatively infrequently in generic training data and are therefore prone to substitution or omission errors. To improve recognition of such entities, we continuously mine domain-specific entity vocabularies and construct corresponding ASR training data through the entity-mining and data-synthesis pipeline described in Section~\ref{subsec:SFT}.

Figure~\ref{fig:entity-recall} reports entity recall on an internal industry-domain evaluation set covering 15 domains, comparing Doubao-ASR, Tencent Hy-ASR-3.0-preview, and Qwen-Audio-3.0-ASR. Qwen-Audio-3.0-ASR achieves the highest entity recall across all evaluated domains. The improvements are particularly pronounced in Software \& IT, where the evaluation contains many low-frequency and rapidly evolving technical entities such as software libraries, frameworks, and API names. Consistent improvements are also observed across domains including Artificial Intelligence, Healthcare, Agriculture, Manufacturing, and Culture, Sports \& Media. Overall, these results demonstrate the effectiveness of Qwen-Audio-3.0-ASR in recognizing long-tail industry-specific entities across a diverse range of domains. The observed gains are also consistent with the expected benefits of the entity-mining and data-synthesis pipeline, although other improvements introduced in Qwen-Audio-3.0-ASR may also contribute to the overall performance gains.

\subsection{Hotword Customization}
\label{subsec:hotword}

Within Qwen-Audio-3.0-ASR, we introduce a hierarchical hotword customization mechanism. Retrieved hotword candidates are divided into two priority levels: a high-priority set (\textbf{P0}) and a broader candidate set (\textbf{P1}). During decoding, the two sets are assigned different conditioning strengths, with P0 receiving a larger weight than P1. This design allows the model to emphasize high-confidence hotwords while still benefiting from a broader set of candidate terms.

Table~\ref{tab:hotword-recall} reports hotword recall with and without hotword conditioning across several long-tail entity categories. With hotword conditioning enabled, Qwen-Audio-3.0-ASR achieves the highest recall on all evaluated P0 categories, demonstrating the effectiveness of the hierarchical hotword mechanism for improving recognition of high-priority long-tail entities.

For subject terms and trending buzzwords, the improvement in P1 recall is smaller than that observed for the corresponding P0 categories, which is consistent with the weaker conditioning weight assigned to P1 candidates.

\begin{table}[h]
\centering
\caption{Hotword recall (\%) without vs.\ with hotword conditioning, by category and priority tier (P0/P1).}
\label{tab:hotword-recall}
\begin{adjustbox}{width=\linewidth}
\begin{tabular}{lcccccc}
\toprule
& \multicolumn{2}{c}{Doubao-ASR} & \multicolumn{2}{c}{Fun-ASR-Flash} & \multicolumn{2}{c}{Qwen-Audio-3.0-ASR} \\
\cmidrule(lr){2-3} \cmidrule(lr){4-5} \cmidrule(lr){6-7}
Category & No hotword & +Hotword & No hotword & +Hotword & No hotword & +Hotword \\
\midrule
Person names (P0)        & 32.27 & 82.64 & 37.33 & 95.26 & 62.12 & \textbf{99.43} \\
Subject terms (P0)       & 75.14 & 93.06 & 71.84 & 92.53 & 69.36 & \textbf{99.42} \\
Subject terms (P1)       & 84.36 & 93.04 & 86.02 & \textbf{96.73} & 88.46 & 95.31 \\
Trending buzzwords (P0)  & 48.24 & 91.95 & 43.26 & 82.53 & 63.08 & \textbf{99.46} \\
Trending buzzwords (P1)  & 51.24 & \textbf{89.48} & 53.92 & 87.00 & 74.00 & 88.34 \\
AI entities (P0)         & 44.21 & 65.24 & 79.88 & 90.55 & 88.22 & \textbf{99.39} \\
Brand names (P0)         & 44.18 & 57.88 & 80.14 & 95.55 & 83.83 & \textbf{97.95} \\
Product entities (P0)    & 67.59 & 78.70 & 73.64 & 89.92 & 72.87 & \textbf{99.07} \\
\bottomrule
\end{tabular}
\end{adjustbox}
\end{table}

\subsection{Long-audio Context Modeling}
\label{subsec:long-audio-context}
Many everyday recognition errors are homophone or first-mention-entity confusions that are only resolvable with context beyond the current utterance -- for example, disambiguating a technical filename from a similar-sounding common phrase, or recognizing a person's name correctly only after it has already appeared earlier in the same conversation. Frame-local acoustic decoding cannot resolve these cases because its receptive field does not extend across utterance boundaries.

Building on the contextual supervised fine-tuning described in Section~\ref{subsec:SFT}, Qwen-Audio-3.0-ASR conditions the transcription of the current audio segment on recently recognized text from the same session, functioning as a running context cache that requires no manually maintained hotword list. 

Table~\ref{tab:long-context-examples} illustrates representative corrections enabled by this mechanism: entities and technical terms that recur across segments (e.g., a codebase name, a repeated technical abbreviation, or a person's name introduced earlier) are transcribed consistently once they are anchored by prior context, whereas the same segment decoded without that context reverts to a phonetically plausible but incorrect alternative.

\begin{table*}[t]
\centering

\begin{CJK*}{UTF8}{gbsn}

\caption{
Representative corrections enabled by long-audio context.
The context cues occurred earlier in the same session.
Colored spans highlight recognition errors and their
context-conditioned corrections.
}
\label{tab:long-context-examples}

\begingroup
\small
\renewcommand{\arraystretch}{1.22}
\setlength{\tabcolsep}{5pt}
\arrayrulecolor{QwenPurple!35}

\begin{tabular}{
  @{}
  V{0.27\textwidth}
  V{0.23\textwidth}
  V{0.22\textwidth}
  V{0.22\textwidth}
  @{}
}
\toprule

\rowcolor{QwenPurple!14}
\textbf{Earlier context cue}
&
\textbf{Reference}
&
\textbf{Without long context}
&
\textbf{With long context}
\\

\midrule

\texttt{xsltproc style.xsl input.xml}
occurred earlier
&
\texttt{style.xsl data.xml > out.html}
&
\texttt{\bad{xsl} data.xml > out.html}
&
\texttt{\good{style.xsl} data.xml > out.html}
\\

\rowcolor{QwenLavender}
Earlier explanation:
“这叫法条竞合”
&
法条竞合
&
\bad{划掉即可}
&
\good{法条竞合}
\\

The term \texttt{unigram}
occurred twice earlier
&
我们称之为 \texttt{unigram}
&
我们称之为 \bad{\texttt{unique}}
&
我们称之为 \good{\texttt{unigram}}
\\

\rowcolor{QwenLavender}
\texttt{TCC} was repeatedly
established as the metric name
&
\texttt{our TCC will be...}
&
\texttt{our \bad{GGC} will be...}
&
\texttt{our \good{TCC} will be...}
\\

The name “洛军是阿占的儿子”
occurred earlier
&
洛军是阿占的儿子
&
\bad{洛君}是阿占的儿子
&
\good{洛军}是阿占的儿子
\\

\rowcolor{QwenLavender}
The technical term
\texttt{softmax} occurred earlier
&
\texttt{softmax} 形式
&
\texttt{\bad{softmap}} 形式
&
\texttt{\good{softmax}} 形式
\\

The mode name
\texttt{debug} occurred earlier
&
只能 \texttt{debug} 模式
&
只能 \bad{第八个} 模式
&
只能 \good{\texttt{debug}} 模式
\\

\rowcolor{QwenLavender}
The name “张贵生”
occurred earlier
&
张贵生
&
\bad{张桂生}
&
\good{张贵生}
\\

\bottomrule
\end{tabular}

\endgroup
\end{CJK*}
\end{table*}

\begin{figure}[h]
    \centering
    \includegraphics[width=1\linewidth]{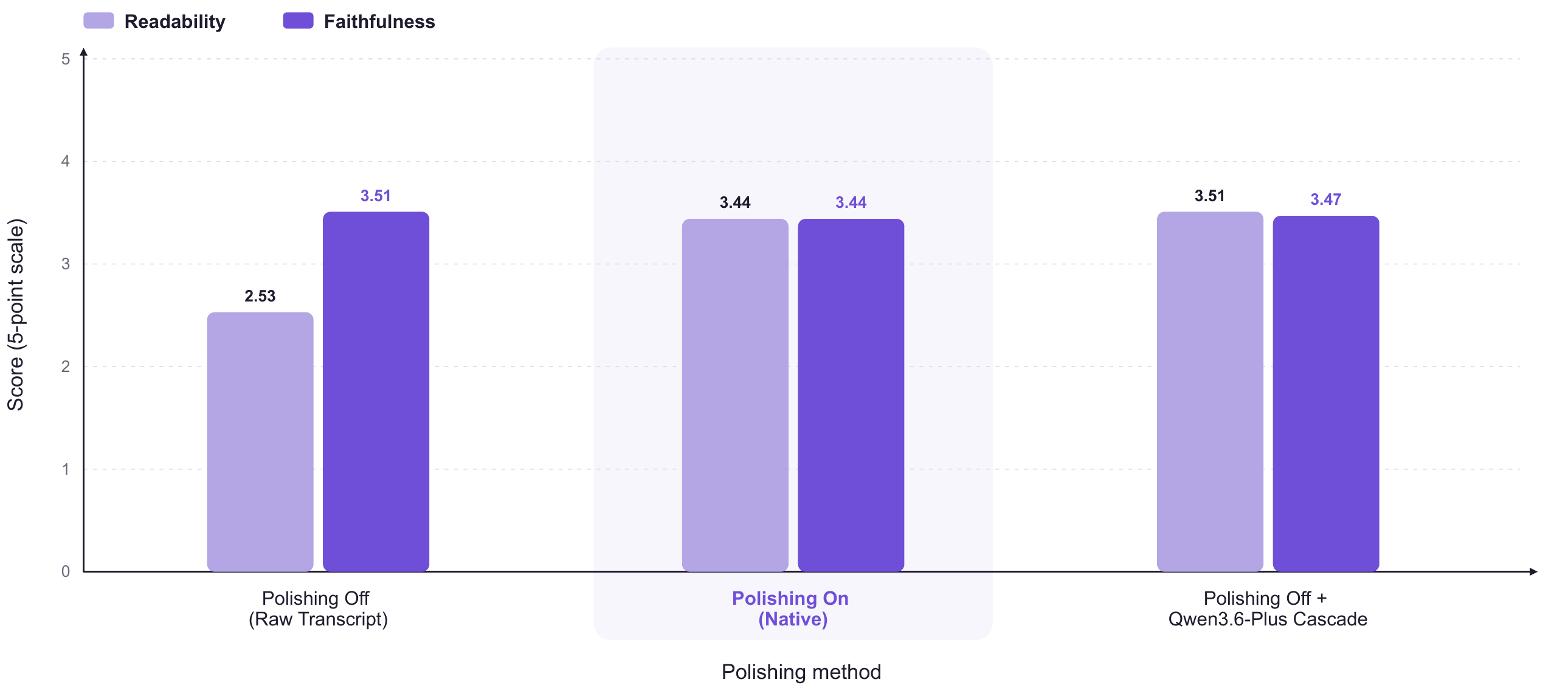}
    \caption{Readability and faithfulness of native polishing vs.\ the raw transcript and vs.\ an external ASR+LLM cascade.}
    \label{fig:polishing}
\end{figure}

\subsection{Native Transcription Polishing}
\label{subsec:native-polishing}
Conventional pipelines address filler words, stutter-like repetitions, and mid-sentence self-corrections through a second, cascaded stage: an LLM rewrites the raw ASR transcript after it is produced. This decouples acoustic decoding from text refinement, but it adds a second autoregressive generation pass whose latency lands directly on the critical path between end-of-utterance and text availability.

Qwen-Audio-3.0-ASR instead performs this cleanup within the same decoding pass as recognition: an instruction flag toggles whether the model emits the raw transcript or a polished version that removes filler words and stutter-like repetitions, resolves self-corrections in favor of the speaker's final intent, and normalizes punctuation and formatting, all without an additional model call. Figure~\ref{fig:polishing} compares this native, single-pass polishing against the same model's raw output and against a conventional two-call cascade (the same raw ASR output rewritten by Qwen3.6-Plus). Native polishing raises the readability score from 2.53 to 3.44 (on an internal 5-point scale), reaching a faithfulness score (3.44) matching the two-call cascade closely (3.47), while requiring only a single inference pass instead of two.

\begin{figure*}[t]
\centering
\includegraphics[width=\textwidth]
{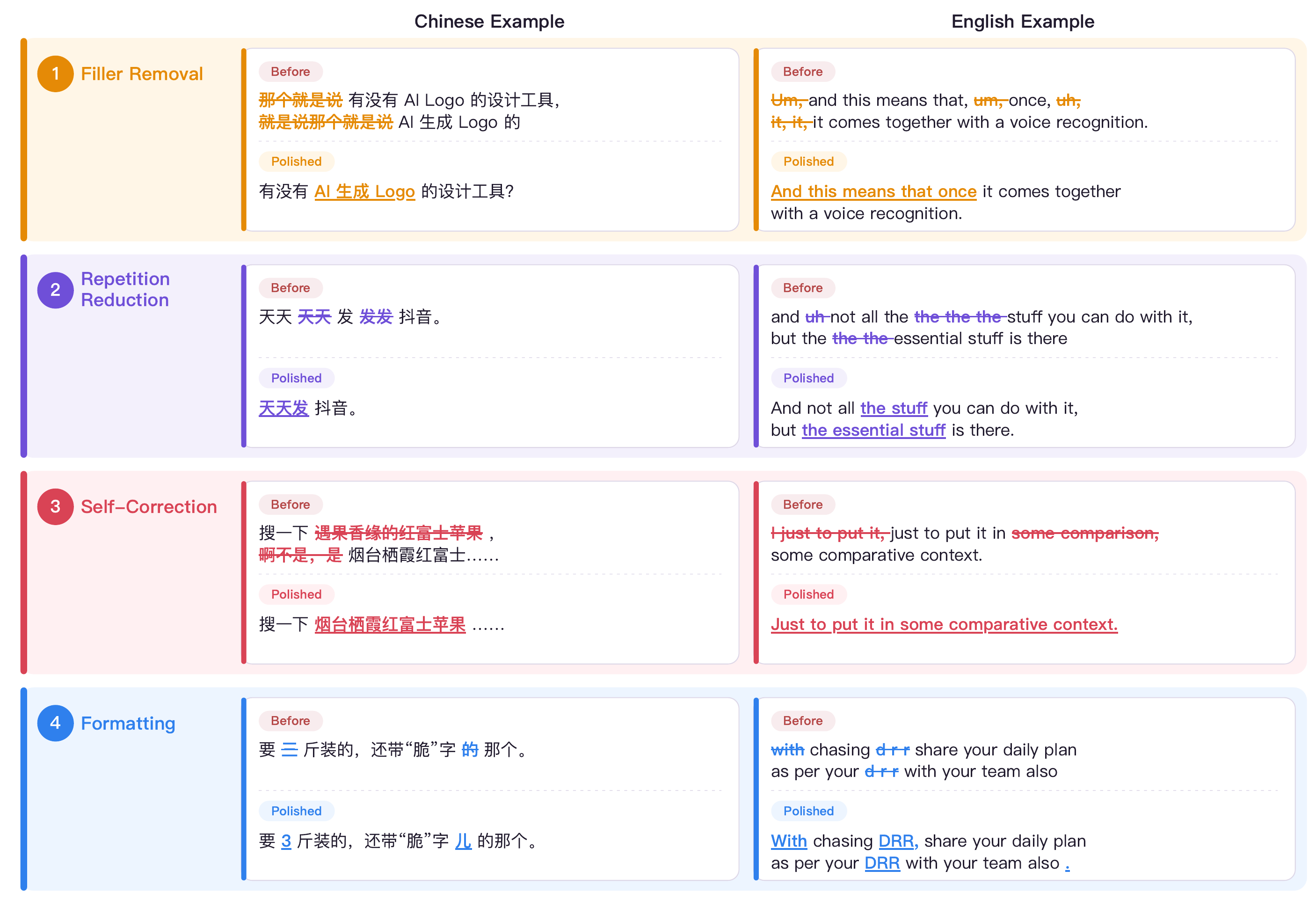}
\caption{
Representative Chinese and English polishing examples.
The model removes fillers and repetitions, resolves
self-corrections, and normalizes sentence structure,
punctuation, and acronyms while preserving the intended meaning.
}
\label{fig:bilingual-polishing-examples}
\end{figure*}

Figure~\ref{fig:bilingual-polishing-examples} presents representative Chinese and English polishing examples. Taking the English cases as an illustration, the model removes hesitation fillers such as \textit{um} and \textit{uh}, collapses repetitions such as \textit{the the the}, and resolves false starts by retaining the speaker's final intended expression. For example, the fragmented phrase \textit{I just to put it / some comparison} is consolidated into \textit{Just to put it in some comparative context}. The model also normalizes capitalization, punctuation, and acronyms, converting the spoken letter sequence \textit{d r r} into \textit{DRR}. These examples show that native polishing produces more fluent and readable transcripts while largely preserving the speaker's original meaning.

\begin{figure}[h]
    \centering
    \includegraphics[width=0.95\linewidth]{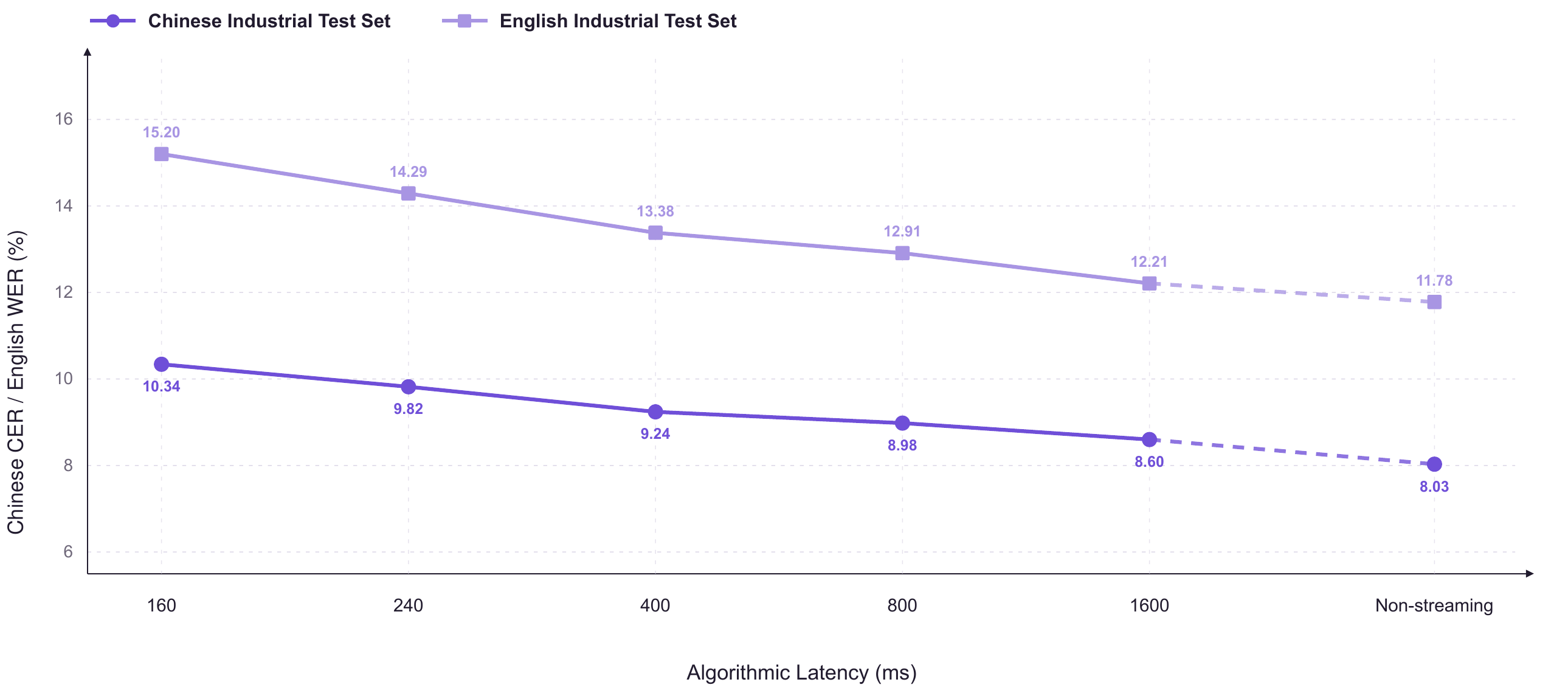}
    \caption{Latency--accuracy trade-off on internal Chinese and English industrial test sets. Chinese CER and English WER decrease as algorithmic latency increases, with the non-streaming setting achieving the lowest error rates.}
    \label{fig:model_latency}
\end{figure}

\subsection{Low-Latency Streaming Recognition}
\label{subsec:streaming-recognition}

Streaming recognition performance is evaluated on internal industrial test sets for Chinese and English. We evaluate multiple streaming configurations with different audio chunk sizes and right-context windows, resulting in the algorithmic-latency operating points shown in Figure 11. All online decoding and latency measurements are performed on an NVIDIA H100 GPU. Both first-token latency and text-display latency are measured from the acoustic end of the corresponding spoken unit to the moment that unit first appears in the on-screen partial transcript. First-token latency applies this definition to the first correctly recognized token displayed on screen in each utterance, whereas text-display latency applies it throughout the utterance. As shown in Figure~\ref{fig:model_latency}, Chinese CER and English WER decrease consistently as algorithmic latency increases, while the non-streaming setting achieves the lowest error rates. These results demonstrate a clear latency–accuracy trade-off and show that the system can be flexibly configured to meet different latency requirements.

Under the low-latency configuration, the system achieves a first-token latency below 200~ms and a text-display latency below 300~ms. At the 400~ms algorithmic-latency operating point in Figure~\ref{fig:model_latency}, the system achieves a Chinese CER of 9.24\% and an English WER of 13.38\%, compared with 8.03\% CER and 11.78\% WER, respectively, in the non-streaming setting. This corresponds to relative error-rate increases of approximately 15.1\% for Chinese and 13.6\% for English.

\section{Conclusion} 
In this report, we present Qwen-Audio-3.0-ASR, a large-scale LLM-based automatic speech recognition system built on the Qwen Mixture-of-Experts backbone and trained on tens of millions of hours of speech data. Through a unified instruction-following framework, Qwen-Audio-3.0-ASR supports transcription across 30 languages and 16 Chinese dialectal varieties, together with production-oriented capabilities including industry-domain entity recognition, hierarchical hotword customization, native single-pass transcription polishing, and long-audio contextual modeling. Extensive evaluations demonstrate strong and well-balanced performance across Chinese, English, multilingual, dialectal, and real-world industrial settings, including substantial improvements in long-tail entity recognition, hotword customization, and transcription readability. In addition to Qwen-Audio-3.0-ASR, we develop Qwen-Audio-3.0-ASR-Streaming as a dedicated low-latency streaming model for interactive applications. We further introduce Message ASR as a deployment profile of Qwen-Audio-3.0-ASR for real-world transcription scenarios, integrating historical context, hierarchical hotwords, native polishing, and low-latency transcription within a unified decoding workflow without introducing a separate recognition or rewriting pipeline. Together, these components provide a comprehensive ASR solution for diverse real-world production requirements.

\section{Authors (in Alphabetical Order of Last Name)}

\begin{multicols}{3}
    \begin{itemize}[noitemsep]
        \item Chuanmeng Bian
        \item Daren Chen
        \item Peixin Chen
        \item Zhigao Chen
        \item Zhiyun Fan
        \item Zhifu Gao
        \item Bo Gong
        \item Qing Gu
        \item Jiajun He
        \item Yawei Hu
        \item Yunjie Ji
        \item Jingbei Li
        \item Xiangang Li
        \item Xu Li
        \item Zengxi Li
        \item Zheng Li
        \item Chengdong Liang
        \item Baiji Liu
        \item Ying Liu
        \item Bin Ma
        \item Yiping Peng
        \item Yuezhang Peng
        \item Zhendong Peng
        \item Yu Pu
        \item Yang Shi
        \item Xin Shu
        \item Jian Tang
        \item Biao Tian
        \item Peiyao Wang
        \item Tianzi Wang
        \item Wen Wang
        \item Wupeng Wang
        \item Cheng Wen
        \item Yuzhong Wu
        \item Zijian Xia
        \item Yunchong Xiao
        \item Nan Yang
        \item Jianwei Yu
        \item Jixing Yu
        \item Binbin Zhang
        \item Lei Zhang
        \item Sitong Zhao
        \item Guangdong Zhou
        \item Yuan Zhou
        \item Jianheng Zhuo
    \end{itemize}
\end{multicols}

\section{Acknowledgments}

We are immensely grateful for the invaluable discussions, support, and assistance provided by many colleagues throughout the development of Qwen-Audio-3.0-ASR. Special thanks go to Qian Chen, Han Zhao, Chong Deng, Chaohong Tan, Cheng Wen, Qinglin Zhang, Xiang Lv, Wei Ju, Luyao Cheng, Haoxu Wang, Yongchao He, Chen Ding, Bangduo Chen, Siyu Wang, Zelan Yang, Changwei Li, and Shengkui Zhao.

\bibliographystyle{unsrtnat}
\bibliography{refs}

\end{document}